\documentclass{isprs} 
\usepackage{subfigure}
\usepackage{setspace}
\usepackage{xcolor}
\usepackage{geometry} 
\usepackage{epstopdf}
\usepackage{multirow}
\usepackage{svg}
\usepackage[labelsep=period]{caption}  
\usepackage[british]{babel} 
\usepackage[para, hang]{footmisc}
\usepackage[colorlinks=true, linkcolor=black, citecolor=black, filecolor=black, urlcolor=black]{hyperref}
\usepackage{natbib}
\usepackage{array}

\makeatletter
\newcommand*{\centerfloat}{%
  \parindent \z@
  \leftskip \z@ \@plus 1fil \@minus \textwidth
  \rightskip\leftskip
  \parfillskip \z@skip}
\makeatother

\newcommand{\turn}[1]{%
  \rotatebox[origin=c]{90}{#1}
  }

\newcolumntype{P}[1]{>{\centering\arraybackslash}p{#1}}
\newcolumntype{M}[1]{>{\centering\arraybackslash}m{#1}}

\newcounter{subsubsub}[subsubsection]

\newcommand{\subsubsubsection}[1]{
    \refstepcounter{subsubsub}
    \subsubsection*{\alph{subsubsub}.~~~~#1}
}

\titlespacing{\subsubsubsection}{0cm}{-0.5cm}{1mm}

\begin{document}

\title{Comparison of two data fusion approaches for land use classification}
\date{31/03/2023}

\author{M. Cubaud\thanks{Corresponding author}, A. Le Bris, L. Jolivet, A-M. Olteanu-Raimond}

\address{Univ Gustave Eiffel, IGN-ENSG, LASTIG, F-94160 Saint-Mandé, France - \\(martin.cubaud, arnaud.le-bris, laurence.jolivet, ana-maria.raimond)@ign.fr}


\commission{XX, }{YY} 
\workinggroup{XX/YY} 
\icwg{}   

\abstract{

Accurate land use maps, describing the territory from an anthropic utilisation point of view, are useful tools for land management and planning. To produce them, the use of optical images alone remains limited. It is therefore necessary to make use of several heterogeneous sources, each carrying complementary or contradictory information due to their imperfections or their different specifications. This study compares two different approaches i.e. a pre-classification and a post-classification fusion approach for combining several sources of spatial data in the context of land use classification. 
The approaches are applied on authoritative land use data located in the Gers department in the south-west of France. Pre-classification fusion, while not explicitly modeling imperfections, has the best final results, reaching an overall accuracy of 97\% and a macro-mean F1 score of 88\%.

}


\maketitle

\section{INTRODUCTION}\label{INTRODUCTION}

    Land Use (LU) describes the socio-economic human activity of an area (\textit{e.g.} agriculture, residential), while Land Cover (LC) describes its physical surface (\textit{e.g.} vegetation, built-up). Land Use and Land Cover (LULC) maps are very useful for understanding, monitoring, planning and predicting the evolution of the territory. 
    There is no direct relation between LU and LC \citep{cihlarLandCoverLand2001} as there can be several uses in an area with the same land cover (\textit{e.g.} residential or commercial uses in built-up areas) and several covers for the same use (\textit{e.g.} garden and houses in a residential area). Since radiometry and texture from imagery are closely related to LC, traditional remote sensing techniques encounter limitations for LU classification. For this reason, several LULC products show a confusion between land use and land cover \citep{comberUsingSemanticsClarify2008}: their nomenclatures sometimes mix LU and LC classes at the same level.
    However, some previous studies have tried to solve this issue 
     using only optical imagery: by learning LC and LU classification simultaneously through iteration \citep{zhangJointDeepLearning2019} or using graph neural networks to learn topological relationships between previously segmented LC 
     areas 
     \citep{liMappingLandUse2020,liuCNNEnhancedHeterogeneousGraph2022}. 
    A map translation approach has also been implemented by \citet{baudouxMultinomenclatureMultiresolutionJoint2023}.\\ Another approach 
    is to consider complementary sources of information, such as imagery from other sensors, LiDAR data, authoritative databases, volunteered geographic information\\ (VGI), or involuntary geographic information (iVGI) \citep{seeCrowdsourcingCitizenScience2016}. For instance, for LU area classification, \citet{tuRegionalMappingEssential2020} used a Random Forest classifier to classify LU from classical optical, night lights intensity and radar imagery, Points of Interest (POI) from Baidu and demographic data from WorldPop.
    \citet{mengDetectResidentialBuildings2012} detected residential buildings by combining images, a Digital Surface Model extracted from LiDAR, and distance from major roads from an authoritative database using a decision tree classifier.
    \citet{liuDataFusionbasedFramework2021} fused VGI from several mapathon campaigns and \textit{in-situ} assessments using the Dempster-Shafer Theory (DST) to classify the use of LC changes.    \citet{panLandUseClassificationUsing2013} used iVGI from Taxi GPS traces to deduce the social function of some places using Support Vector Machine. \citet{heAccurateEstimationProportion2021} combined optical images and user density of the Tencent web application (iVGI) with a convolutional neural network to classify LU area.
    At the feature level, \citet{fonteCLASSIFICATIONBUILDINGFUNCTION2018} identified building functions using a rule based classifications of OpenStreetMap (OSM), Facebook and Foursquare VGI data, individually, whereas 
    \citet{dengIdentifyUrbanBuilding2022} identified building functions from images, POI and building footprint from Gaode map (authoritative database) and distance to OSM roads using a XGBoost classifier.\\
    The fusion process can be done either before or after classification \citep{joshiReviewApplicationOptical2016}. 
    In pre-classification fusion, 
    all the attributes are concatenated and a machine learning algorithm will predict LU classes from all sources simultaneously. The advantage 
    is that the classifier can exploit the joint information of the sources. On the other hand, in post-classifica\-tion fusion, a prediction is made for each source before they are merged to obtain a final prediction. Post-classification fusion has a greater adaptability: it is easier to add a new source. Among the previously cited articles using data fusion, only \citet{fonteCLASSIFICATIONBUILDINGFUNCTION2018} and \citet{liuDataFusionbasedFramework2021} have a post-classification approach, the others having a pre-classification approach.\\
    Moreover, some post-classification fusion algorithms can model the imperfections of the sources and especially the lack of information for some classes according to some sources. Indeed, we believe that the imperfections of the data sources need to be taken into account when combining multiple sources to derive more robustly and precisely LU. Indeed, data sources may have an imperfect internal quality: errors in geometry or attributes, incompleteness, low accuracy due to fuzzy boundaries or to low-level nomenclature. 
    Data sources may have external quality issues if they don't perfectly fit the user's objective.
    It can for instance come from the source being only partly relevant, from the differences of scale of the sources, or from the ambiguities in the meaning of the classes for the different sources \citep{batton-hubertGeographicDataImperfection2019}.\\
    The OCS GE\footnote{\url{https://geoservices.ign.fr/ocsge}} (Large Scale Land Use Land Cover) is a LULC map produced by the French National Mapping Agency (IGN) with separated LU and LC nomenclatures. It partitions the space into non-overlapping polygons and assigns to each of them a unique LU and a unique LC class (Table~\ref{tab:OCSGE_classes}).
    \begin{table}[h]
	\centering
		\begin{tabular}{|c|c|}\hline
			\multicolumn{2}{|c|}{LU Classes}\\\hline
            Code & Description\\\hline
            LU1 & Primary production\\
            LU2 & Secondary production\\
            LU3 & Tertiary production\\
            LU4 & Logistic transport networks and infrastructures\\
            LU5 & Residential use\\
            \multirow{2}{*}{LU6} & Other use (Areas under construction, Abandoned\\
            & areas, Unused areas or Unknown use)\\\hline
            LU235 & Secondary, tertiary or residential use\\\hline
		\end{tabular}
    \caption{Level 1 land use classes of OCS GE.}
    \label{tab:OCSGE_classes}
    \end{table}
    LU is currently assigned to OCS GE polygons through an automatic rule-based process taking as inputs Land Files and topographic information, combined with an intensive manual correction step based on photo-interpretation 
    \citep{ignDescriptionFonctionnelleProcessus2022}. In previous versions of OCS GE, 
    mainly due to lack of accurate information on building use, classes LU2 (secondary production, \textit{i.e.} industrial and manufacturing activities), LU3 (tertiary production, \textit{i.e.} commercial and services activities) and LU5 (residential use) were grouped together into a single class LU235 which limits the calculation of artificialization indicators.\\
    Hence, the aim of this paper is to compare the pre- and post-classification data fusion approaches to distinguish these three LU classes using machine learning techniques.
    Our hypotheses are as follows: (1) Multiple data sources are available and complement each other, (2) and a machine learning model can leverage these sources to infer LU, (3) resulting in improved performance compared to using a single source.\\
    The major contributions of this work are: (1) to propose a general workflow for urban LU classification, (2) to define several attributes from heterogeneous sources to characterize LU polygons, and (3) to compare several approaches and variants to identify the best fusion process for LU classification. Note that our work supposes that the boundaries of LU class exist and are represented by polygons; in the following they are named LU polygons.\\
    The paper is organized as follows. The general proposed methodology and its specific application to distinguish LU235 will first be described (section \ref{METHODOLOGY}). The results will then be presented and compared for both pre- and post-classification approaches (section \ref{RESULTS}), before being discussed in section \ref{DISCUSSION}.

\section{METHODOLOGY}\label{METHODOLOGY}
    \subsection{General Workflow to distinguish land use classes}
\begin{figure}[ht!]
            \begin{center}
        		\includegraphics[width=1.0\columnwidth]{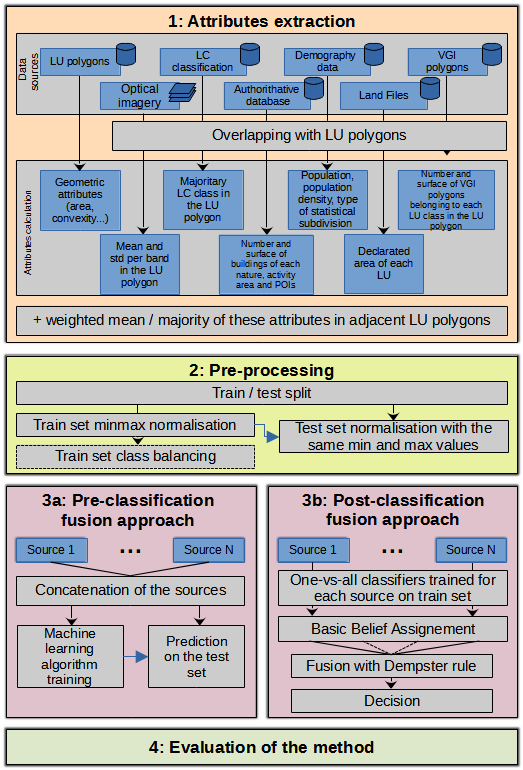}
        	\caption{General workflow for Land Use classification using multiple heterogeneous data sources.}
            \label{fig:workflow}
            \end{center}
        \end{figure}

        The general workflow of the proposed method is illustrated in Figure~\ref{fig:workflow}.
        Steps \ref{construction}, \ref{Preprocessing} and \ref{metrics} are common to both pre- and post-classification approaches.
        The proposed workflow first calculates a set of attributes out of different sources of information. Then a machine learning workflow is trained out on labeled LU data
        to distinguish existing polygons into LU2, LU3 and LU5. Two variants are considered : a first one relying on a single classifier using all available attribute vs. another one involving one classifier per source before these per source results are merged.

        \subsubsection{Attributes extraction:}\label{construction}
        The first step is to extract attributes from the different sources to characterize the LU polygons. Each data source is overlapped with LU polygons, and its data are aggregated to construct attributes at the scale of the LU polygon.\\ To take into account the spatial relationships between the different uses, the mean (or majority value if it is a categorical attribute) of each attribute mentioned above is also computed over the neighboring adjacent LU polygons,
        weighted by the length of the common perimeter.  The weights make it possible to give neighbors a more or less important influence depending on how much border they share. These averages are then used as new attributes. The list of used sources and created attributes are presented in subsection \ref{DataSources}.

        \subsubsection{Pre-processing:}\label{Preprocessing}
            80\% of the dataset is randomly allocated for training, and the remaining for testing.
            Categorical attributes are ordinal-encoded.
            A minmax normalization is applied to each attribute. In order to prevent data leakage, min and max are calculated on the train set and the same values are used for the test set.
            Finally, as there is an important class imbalance, we choose to upsample
            the minority classes (LU2 and LU3) in the train set using the SMOTE-NC algorithm (Synthetic Minority Oversampling Technique for Nominal and Continuous \citep{chawlaSMOTESyntheticMinority2011}). The Python library imblearn \footnote{\url{https://imbalanced-learn.org}} has been used. Other balancing techniques are tested and discussed in subsection \ref{preprocessingEffects}. The test set remains unbalanced to better represent real data and class frequencies.
        \subsubsection{Inference}
        \subsubsubsection{Pre-classification fusion approach:}\label{preclassif}
            A machine learning algorithm is trained on the train set using all attributes from all sources. It is then evaluated on the test set. We compared three widely used machine-learning algorithms: Random-Forest (RF) \citep{breimanRandomForests2001}, Support Vector Machine (SVM) \citep{vapnikSupportVectorMethod1998} and Gradient Boosted Trees (XGBoost) \citep{chenXGBoostScalableTree2016}. 
            The Python libraries Scikit-learn\footnote{\url{https://scikit-learn.org}} and XGBoost\footnote{\url{https://xgboost.readthedocs.io}} were used.
            Most hyperparameters were kept to their default values in the Python libraries, but we tuned via a 5-fold cross-validation some that we considered relevant:
            \textit{n\_estimators} (50, 100, 150, 500, 1000), \textit{max\_attributes} (sqrt, log2, 0.2) and \textit{max\_\linebreak{}depth} (None, 2, 5, 10, 20, 50) for RF, \textit{kernel} (linear, poly or RBF), \textit{C} (1, 10, 100, 1000) and \textit{gamma} (0.001, 0.0001) for SVM, and \textit{n\_estimator} (50, 100, 400, 700, 1000) and \textit{learning\_rate} (0.5, 0.2, 0.1, 0.05, 0.02) for XGBoost.
        \subsubsubsection{Post-classification fusion approach:}\label{postclassif}
            This second approach is based on 
            the Dempster-Shafer theory (DST) \citep{shaferMathematicalTheoryEvidence1976}. 
            The advantages of this framework are its ability to model explicitly uncertainty, imprecision, and incompleteness \citep{olteanu-raimondKnowledgeFormalizationVector2015}.\\
            Let's define the frame of discernment $\Theta$ = \{LU2, LU3, LU5\}, which contains the exhaustive and exclusive hypotheses of our problem. The corresponding referential of definition is the powerset of $\Theta$,
            $2^\Theta$ = \{\{LU2\}, \{LU3\}, \{LU5\}, \{LU2, LU3\}, \{LU2, LU5\}, \{LU3, LU5\}, \{LU2, LU3, LU5\}\}. It doesn't contain the empty set because the closed world assumption is made, \textit{i.e.} our frame of discernment is truly exhaustive.\\
            Each source of information will make for each LU polygon a basic belief assignment (bba) and create a mass of belief $m_s(.) : 2^\Theta \rightarrow [0, 1]$ such that $\sum_{H \in 2^\Theta}m_s(H)=1$.\\
            In order to assign these bba, we trained for each source a one-vs-all XGBoost classifier per singleton hypothesis. Each classifier returns a probability $P_H$ for a hypothesis $H\in \Theta$ and $1 - P_H$ for $\neg H$. The bba is then defined as follows:
            \begin{equation}
                \left\{
                    \begin{array}{cc}
                        m(H) = \frac{P_H}{|\Theta|},~ m(\neg H) = \frac{1 - P_H}{|\Theta|}&\forall H \in \Theta\\
                        m(H) = 0 & \textrm{for all the others,}\\
                    \end{array}
                \right.  
            \end{equation}
            with $|\Theta|$ is the number of singleton hypotheses (here 3). This method to do the bba assignment is inspired by \citep{appriouUncertainDataAggregation1998}. Among other tested method, this one appears to best model the doubts of the source about the hypothesis.\\
            These bba are then merged using Dempster's rule of combination \citep{shaferMathematicalTheoryEvidence1976}. It is a commutative and associative rule of fusion that strengthens the mass of belief for the hypotheses on which the sources agree and that redistributes the conflict $\kappa$ (when the sources believe in incompatible hypotheses) proportionally to the masses.\\
            Once all sources have been fused, the final decision for each LU polygon is made by 
            selecting the hypothesis in the frame of discernment $\Theta$ with the highest pignistic probability \citep{smetsTransferableBeliefModel1994}. An advantage of using pignistic probabilities over other decision functions such as credibility or plausibility is that the results can be interpreted as probabilities, \textit{i.e.} values ranging between 0 and 1 and with a sum for the singleton hypotheses of $\Theta$  equal to 1.

        \subsubsection{Evaluation of the method:}\label{metrics}
        Predicted LU are compared to the ground-truth. 
        As the overall accuracy (OA) 
        can be biased by the high class imbalance, we focused on the macro-mean F1 score (mF1). mF1 gives the same weight for the good classification of each class in terms of both recall and precision. It is constructed from the confusion matrix M and the per class recall (r) and precision (p):
        \begin{equation}
            r_i=\frac{M_{ii}}{\sum_{j=1}^{c}M_{ij}},~~~
            p_i=\frac{M_{ii}}{\sum_{j=1}^{c}M_{ji}},~~~
            F1_i=2\frac{r_i p_i}{r_i+p_i}
        \end{equation}
        \begin{equation}
            OA=\frac{\sum_{i=1}^c M_{ii}}{\sum_{i,j=1}^c M_{ij}},~~~
            mF1=\frac{1}{c}\sum_{i=1}^cF1_i
        \end{equation}
        with c the number of classes and $M_{ij}$ the number of elements of class i in ground truth that are predicted in class j.

    \subsection{Study Area}
\begin{figure}[!ht]
        \begin{center}
    	    \includegraphics[width=1.0\columnwidth]{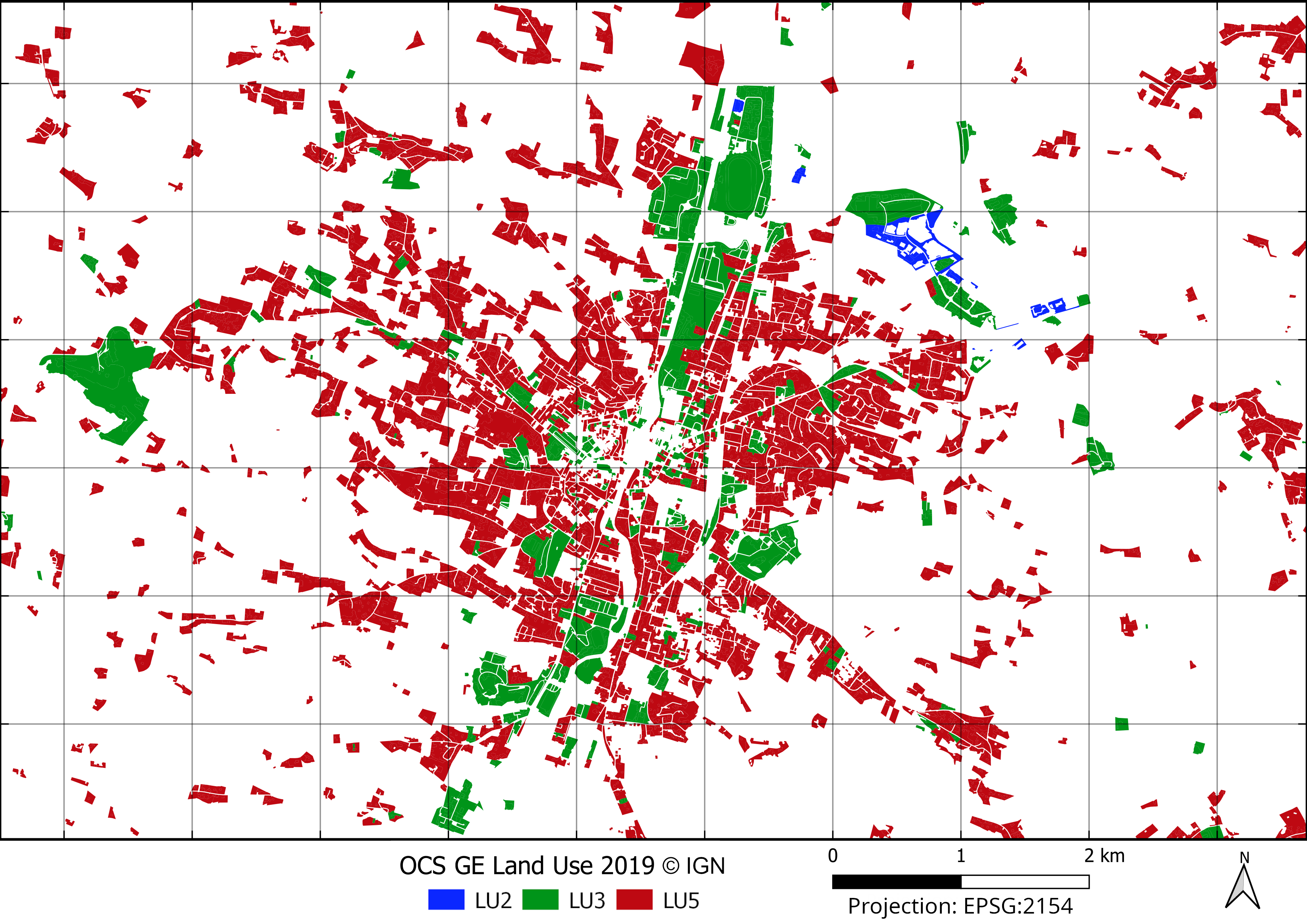}
    	\caption{Example of ground truth over Auch, the largest town of Gers ($\approx~20,000$ inhabitants)}
        \label{fig:GT}
        \end{center}
        \end{figure}

        The 2019 edition of OCS GE in the Gers department, in the South-west of France, 
        has been selected as ground truth. Only LU2, LU3 and LU5 polygons have been kept.
        For each source, its closest available version to 2019 has been used.
        As Gers is mostly rural, with a population density of 33/km², there are few industrial areas and most of the polygons retained are residential. More precisely, in the ground truth, 89.6\% of the 131,224 polygons are LU5, 9.8\% are LU3 and 0.6\% are LU2. Figure~\ref{fig:GT} shows 
        an example of these three classes for the town of Auch.
        According to the data specifications, the accuracy of the polygons' borders is about a meter, and the confusion rate between the classes is less than 5\%.



    \subsection{Data Sources and constructed attributes}\label{DataSources}
        The following subsection presents the used data sources, grouped by types, and discusses their imperfections. The attributes constructed from the sources are defined. Each attribute can provide either explicit information about LU (\textit{e.g.} building functions) or implicit information that may relate indirectly to LU (\textit{e.g.} surface of the LU polygon). A total of 152 attributes has been defined, half of them being neighboring attributes.
        \subsubsection{LU polygons geometry:}
        First, 25 attributes are defined based on the shape of the LU polygon itself, 
        to characterize its geometry. As it is partly linked to LU, these attributes all give implicit information.
        \begin{itemize}
            \item Surface of the polygon
            \item Convexity $=~\frac{area(polygon)}{area(convex\_hull(polygon))}$, which measures the regularity of the LU polygon.
            \item Compactness $=~\frac{4 \pi ~area(polygon)}{perimeter(polygon)^2}$, which compares the shape to a circle.
            \item Elongation $=~\frac{length(OrientedBoundingBox(polygon))}{width(OrientedBoundingBox(polygon))}$, which measures how stretched the polygon is.
            \item Number of holes in the polygon, which is an indicator about if there are smaller LU polygons inside.
            \item Polygonal signature: it characterizes the shape of the polygon; the polygonal signature is a function which gives for each point of the border of the polygon its distance to the center \citep{menerouxRadialDistanceReally2022}. We normalized it by the perimeter of the polygon (so that it is scale invariant) and sampled it with 20 points (so 20 attributes), starting by the closest point to the center, and turning clock-wise. Two polygons with similar shapes (within a scale factor) will have a close signature in the sense of a certain distance called radial distance, the reciprocal being true for a subset of the polygons.
        \end{itemize}

        \subsubsection{Optical imagery:}
        We used the 20~cm resolution open-\linebreak[4]license French national reference ortho-image database (BD ORTHO, 2019). Its planimetric accuracy is 80~cm.
        We computed the eight following attributes for each LU polygon : mean and standard deviation of the blue, green, red and near infrared channels. These attributes provide implicit information.

        \subsubsection{Land Cover:}
        Three land cover maps have been used: OCS~GE (2019) LC class, CORINE Land Cover (CLC, 2012) and OSO (2019), which is an open-license land cover map of France generated yearly \citep{ingladajordiTheiaOSOLand2019} out of Sentinel data. OSO is a raster map, whereas OCS~GE LC and CLC are vector maps. The three maps don't share the same minimal mapping unit: 100~m² for OSO, between 500 and 2500~m² for OCS~GE LC (for built-up and unbuilt areas respectively) and 10000~m² for CLC. CLC is therefore more imprecise. The three maps have an accuracy of about 85\% according to their specifications.
        For each map, we added as attribute the majority land cover class within the LU polygon (one attribute per map).
        OSO and CLC also include some LU classes mixed in their nomenclature, especially they both have an "industrial or commercial area" class. Except for this class, only implicit information about LU is given, as only indirect links exist between LC and LU.
        \subsubsection{Authoritative databases:} BD TOPO (2022) is the open-license French reference vector topographic database. Its planimetric accuracy is 2.5~m. Three layers are selected giving explicit or implicit information about LU: "building",  "Area of activity or interest" and  "Establishments Receiving the Public (ERP)". The building layer has a completeness of 95\%. It indicates a main building function.
        Nevertheless, about 60\% of them have an undifferentiated use. For each of the 8~building function (including undifferentiated) the area and the number of buildings inside the LU polygon are computed. When a building intersects with more than one LU polygon, it is counted for each LU polygon and the area attribute of the corresponding building function is increased by the intersected area for each LU polygon. Another building based-attribute is 
        the average height (implicit information). However, this information was lacking for about 10\% of buildings, so the average only includes the buildings for which the height was given, and was else set to 0. In total, seventeen attributes were defined for buildings.\\
        \begin{itemize}
            \item The "Area of activity or interest" layer describes places with a specific economic activity. It usually represents bigger zones than a LU polygon. However, about 40\% of the objects from this layer have a fictive geometry (\textit{i.e.} they are represented by a 25~$m^2$ square) when no geographical extent has been entered.
        \end{itemize}
        Categories and natures has been mapped to LU2 and LU3, and the intersecting area has been calculated as attribute for each LU polygon (explicit information). 
        The ERP layer provides explicit point information about LU3 so the number of its POI in each LU polygon has also been added as attribute.The last two layers are grouped in a single source named "BD TOPO other" in the following.

        \subsubsection{Demography statistical data:}
        This dataset describes population in 2019 at the IRIS (statistical subdivision of the municipality) spatial scale produced by the National Institute of Statistics and Economic Studies (INSEE). Even if this dataset is too coarse to explicitly distinguish LU5, it may help to detect globally residential area. Three attributes are derived from this dataset: population, population density, and the type of IRIS (\textit{i.e.} housing, business, and other) which roughly indicates how the municipality is divided.

        \subsubsection{Land Files:}
        Land files describe tax parcels.
        For each cadastral parcel, an area per land use class in 2019 was obtained by combining several indicators from Land Files, as described in \citet{rutkowskiQualificationUsageZones2017}. We derived four~attributes: LU2, LU3, and LU5 surfaces, as well as the LU class name with the highest surface (explicit information). Known issues are that it is partly based on declarative data and that it fails for public institutions as they don't pay taxes.

        \subsubsection{VGI databases:}
        OpenStreetMap (accessed in 2022) is a worldwide collaborative mapping project. VGI maps can complete information absent from authoritative databases, however they are known to be more incomplete in less populated areas as there are fewer contributors. For instance, the completeness of OSM buildings is around 80\% for European countries including France \citep{qiZhou_osmcompleteness}. Moreover, the position accuracy is metric since 97\% of building geometry is coming from authoritative data building \citep{leGuilcher}. The thematic completeness of buildings is however low, with only 1\% buildings having a tag value different than "yes". As in \citet{fonteCLASSIFICATIONBUILDINGFUNCTION2018}, we selected some OSM polygons
        and points, and mapped them to LU1, LU2, LU3, LU5 and unknown\footnote{\textit{Cf.} \url{https://github.com/mcubaud/OSM_to_OCS_GE_LU}}.
        We then computed the number and surface for each of these classes for OSM polygons (10 attributes), and the number of LU2, LU3 and LU5 points (3 attributes).

\section{RESULTS}\label{RESULTS}
    This section first presents the quantitative results for the classification of LU2, LU3 and LU5 by the two approaches, and then provides a qualitative analysis of the errors.
    \subsection{Comparison of metrics}
    Table~\ref{tab:global_metrics} compares the global results obtained by the algorithms, while Table~\ref{tab:per_class_metrics} compares the per-class results. Training and testing computing time were measured on a computer with a processor AMD Ryzen 5 5500U with Radeon Graphics 2.10 GHz. Overall Accuracy is not very meaningful, as it can be very high (about 90\%) if the algorithm only predicts LU5. In terms of macro-mean F1 score, XGBoost and Random Forest reach similar scores of 88\% and 86\% respectively.
    SVM performs here worst and slowest, for the tested hyperparameter values.
    In terms of computation time, XGBoost has the shortest training time, and XGBoost and DST predict the test set in less than a second. Testing time is much shorter than training time because training is more computationally expensive, and is done on a much larger set since minority classes are oversampled.
    DST testing time is very short here as we have only three classes in our frame of discernment, but DST computational complexity increases exponentially with the number of classes.
    \begin{table}[h]
	\centerfloat
    \resizebox{1.1\columnwidth}{!}{
		\begin{tabular}{|l|l|c|c|c|c|}\hline
			Approach&Algorithm&OA&mF1&Training time&Testing time\\\hline
            \multirow{3}{*}{Pre-classif}&RF&\textbf{97}\%&86\%&2 h 32 min&7.2 s\\
            &SVM&85\%&65\%&3 h 6 min&12 min 59 s\\
            &XGBoost&\textbf{97\%}&\textbf{88\%}&\textbf{42 m}&0.4 s\\
            \hline
            Post-classif&DST&94\%&72\%&1 h 43 min&\textbf{0.3 s}\\
            \hline
		\end{tabular}
    }
	\caption{Global metrics for the variants of the two approaches.}
    \label{tab:global_metrics}
    \end{table}

    \begin{table}[h]
	\centerfloat
    \resizebox{1.15\columnwidth}{!}{
		\begin{tabular}{|l|l|c|c|c|c|c|c|}\hline
			Approach&Algorithm&$r_{LU2}$&$r_{LU3}$&$r_{LU5}$&$p_{LU2}$&$p_{LU3}$&$p_{LU5}$\\\hline
            \multirow{3}{*}{Pre-classif}&RF&67\%&\textbf{84\%}&98\%&83\%&81\%&\textbf{98\%}\\
            &SVM&66\%&49\%&89\%&62\%&33\%&94\%\\
            &XGBoost&\textbf{78\%}&82\%&\textbf{99\%}&81\%&\textbf{89\%}&\textbf{98\%}\\
            \hline
            Post-classif&DST&31\%&74\%&97\%&\textbf{88}\%&73\%&97\%\\
            \hline
		\end{tabular}
    }
	\caption{Per-class metrics for the variants of the two approaches. r: recall, p: precision, LU2: secondary production,\\ LU3: tertiary production, LU5: residential use.}
    \label{tab:per_class_metrics}
    \end{table}

    Looking more closely at the per class results, even though we have balanced our train set, the minority class LU2 still has a lower recall for all the algorithms except SVM. The majority class LU5 is the best predicted in terms of recall and precision.\\
    Table~\ref{tab:conf_matrix} gives the confusion matrix obtained by the best algorithm, XGBoost. 
    For all algorithms, the number of errors between LU2 and LU5 is lower than the number of errors between LU2 and LU3 or LU3 and LU5.
    \begin{table}[!h]
	\centering
		\begin{tabular}{|l|ccc|}\hline
			\multirow{2}{*}{Ground Truth}&\multicolumn{3}{c|}{Predicted}\\
            \cline{2-4}
            &LU2&LU3&LU5\\
            \hline
            LU2&131&27&11\\
            LU3&24&2120&431\\
            LU5&7&238&23256\\
            \hline
		\end{tabular}
	\caption{Confusion matrix for XGBoost.}
    \label{tab:conf_matrix}
    \end{table}

    \subsection{Analysis of errors}\label{errors}
\begin{figure}[!ht]
    \begin{center}
		\includegraphics[width=1.\columnwidth]{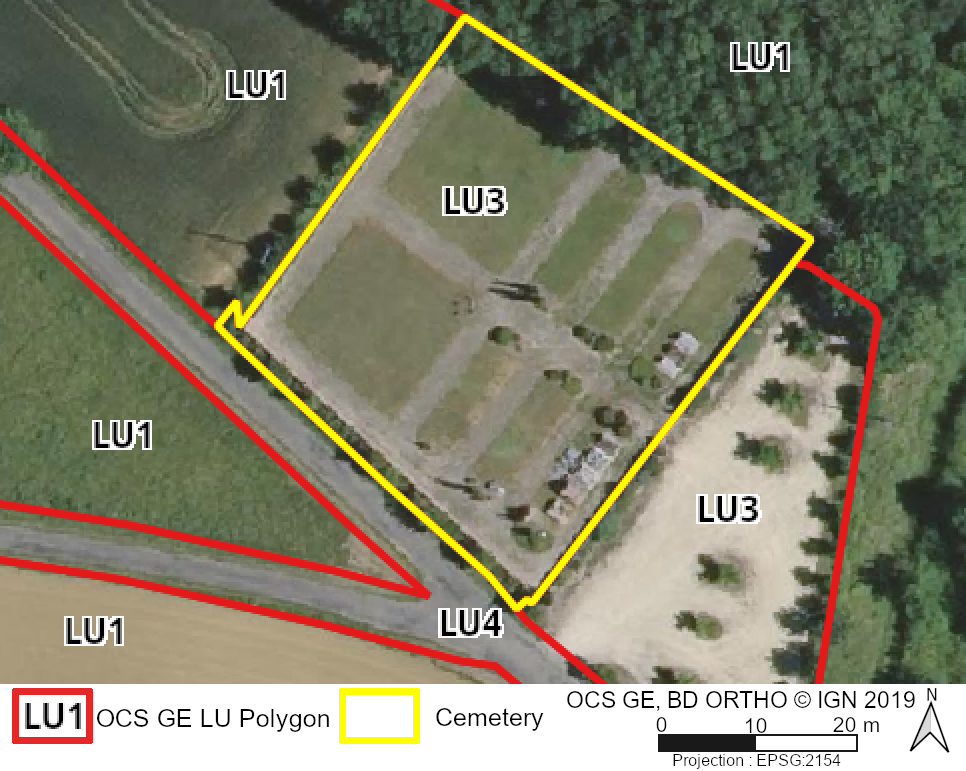}
	\caption{Example of a LU3 polygon with no explicit information in the data sources, predicted LU5.}
    \label{fig:cemetery}
    \end{center}
    \end{figure}

\begin{figure}[!ht]
    \begin{center}
		\includegraphics[width=1.\columnwidth]{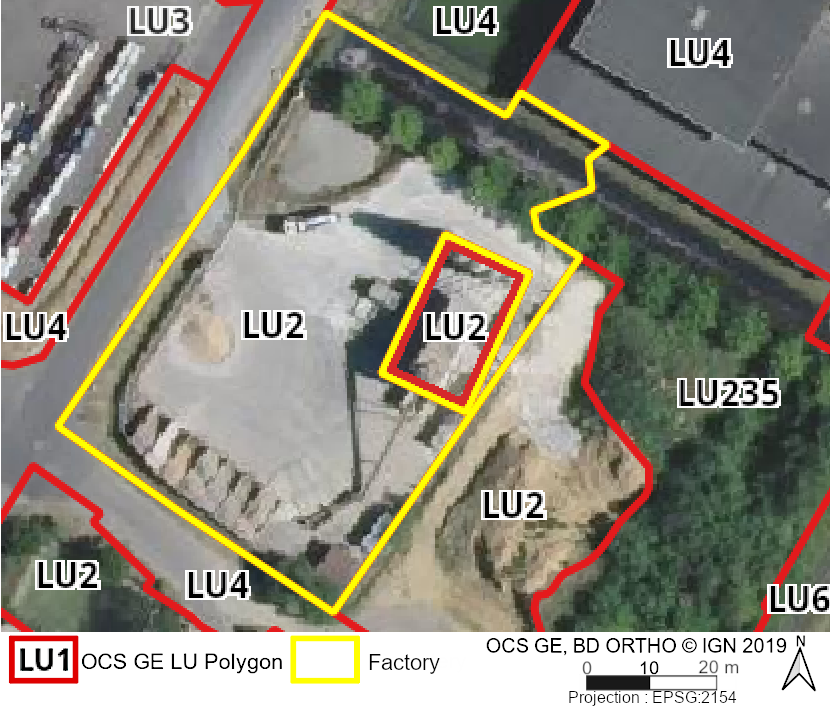}
	\caption{Example of LU2 polygon with incorrect information in the data sources, predicted LU3.}
    \label{fig:concrete}
    \end{center}
    \end{figure}

    This subsection analyses various types and causes of errors. They are illustrated with errors found with XGBoost, although similar errors have also been found with the other algorithms.
     Errors between LU2 and LU3 are mostly encountered in peri-urban areas grouping these two uses that results from local government zoning policies
     , while errors between LU3 and LU5 can sometimes be found in dense town centers where land uses are mixed, and also spatially scattered.\\
    The classifier tends to over-predict LU5 in case no source pro\-vides 
    explicit information about LU. This may be because a source is incomplete 
    or because no source maps this type of element. For instance, the cemetery in Figure~\ref{fig:cemetery} (in yellow) is LU3 in the ground truth but was not present in OSM, nor in the selected layers of BD TOPO, it was thus predicted LU5.
    \\
     Another reason for prediction errors are errors in a source. For example, Figure~\ref{fig:concrete} shows a concrete factory which is LU2 in the ground truth, but it has a commercial use in Land Files, thus it has been predicted LU3.\\
     There can also be a geometric overlay between a LU polygon and another object (\textit{e.g.} a building) represented in a source, which would give the classifier incorrect information.

\begin{figure}[!ht]
    \begin{center}
		\includegraphics[width=1.\columnwidth]{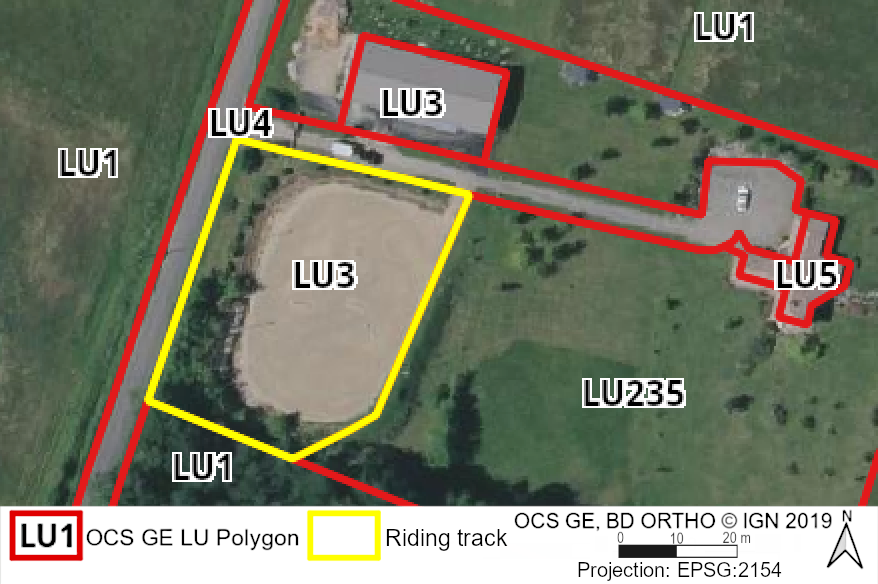}
    \vspace{-0.2cm}
	\caption{Example of a LU3 polygon separated from the object with the relevant information, predicted LU5.}
    \vspace{-0.2cm}
    \label{fig:cheval_genial}
    \end{center}
    \end{figure}

\begin{figure}[!ht]
    \begin{center}
		\includegraphics[width=1.\columnwidth]{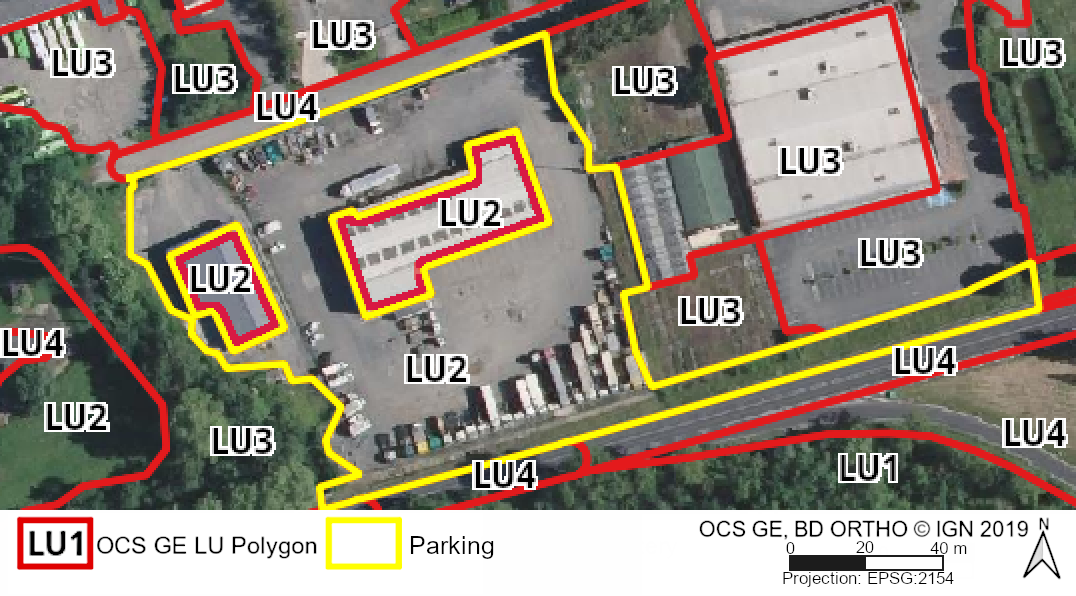}
    \vspace{-0.2cm}
	\caption{Example of a LU2 polygon where neighbors majority is not the most relevant, predicted LU3.}
    \vspace{-0.2cm}
    \label{fig:bad_voisins}
    \end{center}
    \end{figure}

    Using the mean of the attributes over the adjacent LU polygons 
    allows to partially compensate the lack of information. However, it is limited by two elements:
    \vspace{-3mm}
    \begin{itemize}
        \item Sometimes the LU polygon is separated from what should give it the information.
        For instance in Figure~\ref{fig:cheval_genial} a horse riding track, which is LU3 in the ground truth, has no implicit information and is separated from the stable (for which Land Files give the LU3 information) by a road, and was thus predicted LU5.
        \vspace{-2mm}
        \item Sometimes the majority neighborhood is not the most relevant (especially the inner neighborhood should have more weight). For example, in Figure~\ref{fig:bad_voisins}, the buildings and its parking are LU2 in ground truth. The buildings were well predicted thanks to BD TOPO, but the parking has no explicit information. As most neighboring polygons are LU3 (in Land Files attributes), it was predicted LU3.
    \end{itemize}
    \vspace{-3mm}
    As for intrinsic attributes, geometry shifts can also strongly affect these neighbor attributes. For instance, if an object from a source overlaps an OCS GE road polygon, which can be very long, it can give wrong neighborhood information to LU polygons much further away.
    
    \subsection{Contribution of the different sources}\label{contributions_of_sources}
        This subsection investigates several methods to assess the impact of each source on the overall classification.
        Table~\ref{tab:accuracy one source} shows the scores obtained by a XGBoost classifier using only the attributes from one source at a time (including neighbor attributes). No single source achieves high classification performances, it thus justifies the data-fusion process. 
        Using several sources enabled a better classification, especially for LU2. Land Files and layers from BD TOPO obtained the highest scores, which comes from the fact that they carry more information about land use but also may come that they were partly used to construct the ground-truth, which can be seen as a limit of our study. OpenStreetMap is thematically incomplete in Gers and therefore gives the lowest results.

        \begin{table}[!h]
    	\centerfloat
    		\begin{tabular}{|p{2.5cm}|c|c|c|c|c|}\hline
    			Source&OA&mF1&$F1_{LU2}$&$F1_{LU3}$&$F1_{LU5}$\\\hline
                Geometry         &82\%&41\%&~8\%&26\%&89\%\\
                Radiometry       &83\%&46\%&10\%&39\%&90\%\\
                OCS GE LC        &76\%&35\%&~5\%&11\%&89\%\\
                CLC              &72\%&37\%&~5\%&23\%&84\%\\
                OSO              &73\%&33\%&~4\%&~9\%&86\%\\
                BD TOPO building &91\%&63\%&34\%&59\%&95\%\\
                BD TOPO other    &87\%&53\%&21\%&45\%&93\%\\
                Demography       &65\%&36\%&~6\%&23\%&78\%\\
                Land Files       &89\%&57\%&13\%&64\%&95\%\\
                OSM              &29\%&21\%&2\%&17\%&45\%\\
                \hline
                All sources      &97\%&88\%&79\%&85\%&99\%\\
                \hline
    		\end{tabular}
        \vspace{-0.2cm}
    	\caption{Metrics for XGBoost trained only with one source.}
        \vspace{-0.2cm}
        \label{tab:accuracy one source}
        \end{table}

\begin{figure}[!ht]
        \begin{center}
            \includegraphics[width=1.\columnwidth]{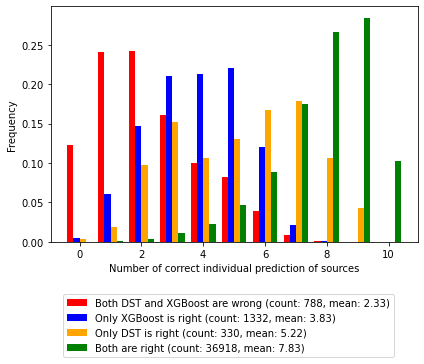}
        \vspace{-0.6cm}
        \caption{Normalized histogram of the frequency of the number of correct individual source prediction according to whether the overall predictions of XGBoost or DST are correct.}
        \vspace{-0.2cm}
        \label{fig:bars}
        \end{center}
        \end{figure}

        LOCO (Leave-One-Covariate-Out) is an attribute importance metric which measures how much score is lost when the model is trained with one attribute dropped \citep{leiDistributionFreePredictiveInference2018}. Table~\ref{tab:loco} shows it for XGBoost when all the attributes from a source are dropped. 
        \begin{table}[!h]
    	\centerfloat
        \resizebox{1.1\columnwidth}{!}{
    		\begin{tabular}{|c|p{2.5cm}|c|c|c|c|c|}\hline
    			&Source dropped&OA&mF1&$F1_{LU2}$&$F1_{LU3}$&$F1_{LU5}$\\\hline
                \multirow{10}{*}{1}&Geometry&0\%&~1\%&~1\%&~0\%&0\%\\
                &Radiometry&0\%&~1\%&~4\%&~0\%&0\%\\
                &OCS GE LC&0\%&-2\%&-5\%&-2\%&0\%\\
                &CLC&0\%&-1\%&-3\%&-1\%&0\%\\
                &OSO&0\%&1\%&~2\%&-1\%&0\%\\
                &BD TOPO building&0\%&~1\%&~1\%&~1\%&1\%\\
                &BD TOPO other&0\%&~8\%&22\%&~0\%&0\%\\
                &Demography&0\%&~1\%&~1\%&~0\%&0\%\\
                &Land Files&2\%&~4\%&~3\%&~8\%&1\%\\
                &OSM&0\%&1\%&~1\%&-1\%&~0\%\\
                \hline
                2&Without neighbors mean&0\%&~4\%&~9\%&~0\%&1\%\\
                \hline
                3&OCS GE LC and CLC&0\%&0\%&-1\%&-1\%&0\%\\
                \hline

    		\end{tabular}
        }
        \vspace{-0.2cm}
    	\caption{Score lost when trained without all the attribute from one source. Negative score lost means that classification actually improves without the source.}
        \vspace{-0.2cm}
        \label{tab:loco}
        \end{table}
        It allows to see that most of the time dropping a source doesn't impact that much classification quality because information may be redundant between several sources. Each source provides new information for only a few LU polygons. BD TOPO's ”Area of activity or interest” layer significantly improves the classification of industrial sites, and Land files the classification of LU2 and LU3. However, as already mentioned, these sources were used for the creation of ground truth. Using information from the neighbor LU polygons is also one of the most important contributors to the final result (part 2 of Table~\ref{tab:loco}). On the contrary, OCS GE LC and CLC seem to worsen the distinction between LU2 and LU3. When both OCS GE LC and CLC are dropped, the improvement in the score is relatively smaller, so they must carry a common part of useful information (part 3 of Table~\ref{tab:loco}).\\
        To explain the differences in the results of the two approaches, Figure~\ref{fig:bars} compares the cases where XGBoost (as the best representative of approach 1) and DST are wrong or right, in terms of the number of sources for which the individual prediction is correct. Most of the time, most of the sources are correct and both XGBoost and DST perform well. When only a few sources are correct, XGBoost often outperforms DST because it doesn't rely on these individual predictions. However, it can sometimes fail even though a majority of sources are correct, which is not possible with DST.

\section{DISCUSSION}\label{DISCUSSION}

    \subsection{Conflict in DST}
        Conflict in DST (the $\kappa = \sum_{X \cap Y = \emptyset}m_1(X)m_2(Y)$ term in the denominator of Dempster's rule 
        ) represents how incompatible the belief masses of each source are and thus quantifies how contradictory the sources are between them.
        Table~\ref{tab:conflict} shows the mean conflict between the bba of each pair of sources, averaged on all test set polygons.
        \begin{table}[!h]
        \centerfloat
        \resizebox{1.1\columnwidth}{!}{
    		\begin{tabular}{|p{2.5cm}|c|c|c|c|c|c|c|c|c|c|}\hline
    			&\turn{Geometry}&\turn{Radiometry}&\turn{OCS GE LC}&\turn{CLC}&\turn{OSO}&\turn{BD TOPO building}&\turn{BD TOPO other}&\turn{Demography}&\turn{Land Files}&\turn{OSM}\\\hline
                Geometry            &08\%&14\%&22\%&25\%&22\%&10\%&14\%&17\%&10\%&24\%\\
                Radiometry          &14\%&13\%&23\%&26\%&23\%&13\%&16\%&19\%&13\%&25\%\\
                OCS GE LC           &22\%&23\%&25\%&29\%&27\%&21\%&23\%&25\%&21\%&29\%\\
                CLC                 &25\%&26\%&29\%&29\%&29\%&25\%&25\%&27\%&25\%&30\%\\
                OSO                 &22\%&23\%&27\%&29\%&25\%&21\%&23\%&25\%&21\%&28\%\\
                BD TOPO building    &10\%&23\%&21\%&25\%&21\%&~5\%&12\%&17\%&~6\%&23\%\\
                BD TOPO other       &14\%&16\%&23\%&25\%&23\%&12\%&13\%&17\%&12\%&24\%\\
                Demography          &17\%&19\%&25\%&27\%&25\%&17\%&17\%&17\%&17\%&26\%\\
                Land Files          &10\%&13\%&21\%&25\%&21\%&~6\%&12\%&17\%&~3\%&23\%\\
                OSM                 &24\%&25\%&29\%&30\%&28\%&23\%&24\%&26\%&23\%&27\%\\
                \hline
    		\end{tabular}
        }
        \vspace{-0.2cm}
    	\caption{Conflict between the different sources.}
        \vspace{-0.2cm}
        \label{tab:conflict}
        \end{table}
    There can be an internal conflict when trying to combine a source with itself, which comes from the fact that Dempster's rule is not idempotent: there is a kind of contradiction in attributing masses to different non-intersecting hypotheses of the referential of definition $2^\Theta$. 
    The three land cover sources (\textit{i.e.} OCS GE LC, OSO and CLC) and OSM are the most highly conflicting sources, which can be explained by the fact that their bba is less accurate. On the contrary, BD TOPO's building layer and Land Files are the least conflicting.
    \subsection{Effects of preprocessing}\label{preprocessingEffects}
    Table~\ref{tab:preprocessing} compares the results obtained by XGBoost, RF or DST when SMOTE-NC, random undersampling (RUS) or nothing are applied to balance the train set. A focus is made on the minority class LU2.\\
    \begin{table}[!h]
        \centerfloat
        \resizebox{1.1\columnwidth}{!}{
    		\begin{tabular}{|c|p{2cm}|c|c|P{1cm}|c|c|}\hline
    			& Sampling strategy & OA & mF1 & Training time & $r_{LU2}$ & $p_{LU2}$\\\hline
                \multirow{3}{*}{\turn{\scriptsize{XGBoost}}}& No balancing & 97\% & 88\% & 12~min & 73\% & 87\%\\
                & RUS & 90\% & 68\% & 6~s & 86\% & 31\%\\
                & SMOTE-NC & 97\% & 88\% & 42~min & 78\% & 81\% \\
                \hline
                \multirow{3}{*}{\turn{RF}}& No balancing & 97\% & 85\% & 1.43~min & 62\% & 87\%\\
                & RUS & 91\% & 72\% & 19 s & 86\% & 41\%\\
                & SMOTE-NC & 97\% & 86\% & 2.5~h & 69\% & 83\% \\
                \hline
                \multirow{3}{*}{\turn{DST}}& No balancing & 90\% & 35\% & 1.6~h & 0 & 0\\
                & RUS & 87\% & 60\% & 15 min & 67\% & 18\%\\
                & SMOTE-NC & 94\% & 72\% & 1.7~h & 31\% & 88\% \\
                \hline
                \multirow{3}{*}{\turn{SVM}}& No balancing & 89\% & 31\% & 16~min & 0 & 0\\
                & RUS & 88\% & 62\% & 1 min & 60\% & 62\%\\
                & SMOTE-NC & 85\% & 65\% & 3.1~h & 66\% & 62\% \\
                \hline
    		\end{tabular}
        }
        \vspace{-0.2cm}
    	\caption{Effects of different balancing of the training set.}
        \vspace{-0.2cm}
        \label{tab:preprocessing}
    \end{table}
    While not balancing tends to under-predict the minority class and random undersampling to over-predict it, SMOTE-NC produces more balanced results. This effect is less significant for XGBoost. For SVM and DST without balancing, no LU2 were predicted at all: because of the high class imbalance, the probability for each classifier to predict LU2 is very low and thus so the final prediction.

    \subsection{Criticism of the ground truth and the data sources}
    Some classification errors can arise from imperfections of the ground truth. They result from the own imperfections of the sources used to constitute the ground truth and from errors made during the rule-based process or by the photo-interpreter. 
    Systematic errors may have been learned by the machine learning process. For instance, according to specifications, repair shops, including garages, should be LU3 as they are considered as a service but are always represented by LU2 polygons in the ground truth and so in the classifier prediction.\\
    According to OCS~GE specifications, cartographic generalization is applied to the polygons in the ground truth. For example, buildings closer than 10 m are aggregated. This triggers more geometry differences between the sources. Moreover, since the shape of the LU polygons no longer matches the image, it makes radiometric and geometric attributes less relevant.\\
    Furthermore, in the current version of OCS GE, there are still some LU235 polygons that are currently supposed to represent mixed land use between the three classes. However, they often still have the old meaning of the LU235 class, \textit{i.e.} the polygon has a single main land use, included in LU2, LU3 or LU5. On the contrary, some LU2, LU3 or LU5 polygons in the ground truth may have a mixed land use.\\
    The quality (\textit{e.g.} semantic and positional accuracy, completeness, actuality, resolution, MMU) of the input data can have an impact on LU classification. In our study, we used both VGI and authoritative data. For the latter, quality is well documented in specifications and metadata files (see section \ref{DataSources}), although some errors and incompleteness may remain. As far as VGI is concerned, of the various sources we identified (\textit{e.g.} Facebook, Foursquare, OSM), only OSM data was finally selected on the basis of the data quality analysis provided by the literature and our qualitative analysis of the area tested. For example, Foursquare is rich in thematic information, but an initial location quality analysis we carried out showed very low position accuracy. Finally, the analysis of the contribution of different sources (see section \ref{contributions_of_sources}) can give a hint on data quality and helps to decide which data sources can ultimately be used.

    \subsection{Comparison with existing works}
    
    The article by \citet{tuRegionalMappingEssential2020} is the closest to this work in terms of objective and nature of the classified objects. However, we couldn't directly compare to its work due to the unavailability of some sources. Comparing between datasets despite the obvious limitations involved, the results for all the metrics are in the same order of magnitude. Our study however benefited from a larger test set, allowing for more robust evaluation. Moreover, it introduced novel aspects such as the comparison of pre- and post-classification fusion approaches and the assessment of the contributions and limitations of individual data sources.

\section{CONCLUSION}\label{CONCLUSION}

Crossing several data sources appears as necessary for an accurate land use classification, however each source can have imperfections that can affect the classification process. Through this article, two data-fusion approaches for land use classification have been compared. After having gathered attributes from the different sources, in the first approach learning is performed using all attributes from all sources at the same time, while in the second approach each source is classified independently and a final prediction is given using the Dempster-Shafer Theory framework.\\
Using a XGBoost classifier, an overall accuracy of 97\% and a macro-mean F1-score of 88\% were obtained. The important class imbalance has been partially resolved by using the SMOTE-NC upsampling algorithm in the train test, but the minority class is still less well classified. The imperfections of each source and their contribution to the final results have been analyzed.\\
Several limitations and perspectives have been identified.
Firstly, this study was limited to industrial, commercial and residential uses, considering that other uses were already well classified, but an interesting perspective is to extend the compared methods to predicting the other classes.
Secondly, the generalization capabilities must be assessed by transferring the trained model to another area, \textit{e.g.} a more urban one. Difficulties may rise from differences in the data sources between the two areas.
Thirdly, another issue is to be able to detect polygons with mixed land uses \citep{nabilInfluenceMixedLanduse2015}, split them if they are in a horizontal mix (the two uses are in distinct areas of the same polygon), and give a mixed land use class in case of vertical mix (\textit{e.g.} an apartment above a shop).\\
Finally, other approaches must be compared in future works. Interesting computer vision techniques such as convolutional neural network would have had here additional difficulties learning the generalized representation.
Building a new ground truth, more accurate and closer to the images, is thus a necessary step for this work.
As shown, using spatial context is useful, and so graph neural networks seem to be promising algorithms for land use classification \citep{liMappingLandUse2020, liuCNNEnhancedHeterogeneousGraph2022}. However, as far as we know, no attempt has been made to use them in this context with more sources than optical images and land cover maps yet.


\setlength{\bibsep}{1.3pt}
\renewcommand\refname{\textbf{REFERENCES}} 
{
	\begin{spacing}{0}
	    \small
		\bibliography{biblio} 
	\end{spacing}
}

\end{document}